\documentclass[11pt]{article}
\usepackage[margin=1in]{geometry}
\usepackage{amsmath,amssymb,amsthm}
\usepackage{bm}
\usepackage{graphicx}
\usepackage{booktabs}
\usepackage{array}
\usepackage{hyperref}
\usepackage{multirow}

\newcommand{\norm}[1]{\lVert #1 \rVert}
\newcommand{\R}{\mathbb{R}}

\newcommand{\Obs}{\mathcal{O}}
\newcommand{\ROF}{\mathrm{ROF}}
\newcommand{\RCF}{\mathrm{RCF}}

\newtheorem{theorem}{Theorem}
\newtheorem{corollary}{Corollary}
\theoremstyle{remark}
\newtheorem{remark}{Remark}

\title{Reward Observability and the Limits of Offline Checkpoint Selection in RSSM World Models}

\author{Nikolai Smolyanskiy\thanks{Code and experimental data:
\url{https://github.com/nsmoly/LunarLander_RSSM} and \newline
\hspace*{1.5em}\url{https://github.com/jonstraveladventures/rof-reproduction}. \newline
\hspace*{1.5em}Video: \url{https://youtu.be/BlRAI2hdKDs}.}\\
\small NVIDIA
\and
Jonathan Shock\\
\small Department of Mathematics and Applied Mathematics, University of Cape Town}
\date{}

\begin{document}
\maketitle
\begingroup
\renewcommand{\thefootnote}{}
\footnotetext{\textbf{Note on this revision.} Versions 1 and 2 claimed that ROF, and a composite score built on it (CROF), predict closed-loop performance and can select checkpoints offline. \textbf{We withdraw those claims.} Collapse varies between training runs with the same data and configuration, and neither ROF nor any other offline statistic we tested predicts it (Section~\ref{sec:negative}). Those versions also reported that the model-based A2C policy beat model-free A2C; scored on 100 held-out episodes, it matches it (Section~\ref{sec:data-efficiency}). This version keeps the world model and its control results, states the Fisher-support theorem (Section~\ref{sec:rof}), and reports which ROF results survive a change of latent coordinates (Section~\ref{sec:validity}).}
\endgroup

\begin{abstract}
We study the closed-loop properties of a recurrent state-space model (RSSM) world model trained on human demonstrations in Gymnasium's LunarLander-v3. We use the trained world model for zero-shot CEM model-predictive control (MPC) and for actor--critic (A2C) training in imagination. Scored on $100$ held-out episodes, the selected model-based A2C policy (trained on world-model checkpoint $280$) reaches a mean return of $+189.5$, matching the best model-free A2C checkpoint ($+183.7$; $600$- and $1000$-step episode caps respectively) with ${\sim}65\times$ fewer real training transitions. We also compare world-model MPC with a behaviour-cloning (BC) policy trained on the successful demonstrations. The BC policy matches MPC's mean return only under stochastic action selection, and on the same $20$ episodes it has one catastrophic episode where MPC has none. We then introduce the Reward Observability Fraction (ROF), the Euclidean fraction of the reward gradient in the observable subspace of the linearized latent dynamics, and show that the next $H$ observations carry Fisher information about every direction in this subspace and none about directions orthogonal to it. ROF itself however depends on how the latent is scaled: rescaling it changes ROF but not the model, so raw levels are not comparable across models. Forcing the reward head onto the posterior-corrected latent $\bm{z}$ raises ROF, and the rise survives a coordinate-invariant check on the pair of runs we tested. Finally, we test whether ROF or other offline metrics can predict the closed-loop collapse of MPC. Collapse varies between training runs with identical data and configuration. ROF does not predict it, none of the $108$ offline summaries we screened passes a permutation test, and the best candidate fails on new runs. Predicting collapse offline from the model and logged data alone remains open.
\end{abstract}

\section{Introduction}
\label{sec:intro}

Model-based reinforcement learning promises to reduce
the number of real-environment interactions needed to learn effective policies
by building a predictive model of the environment, a \emph{world model}, and
using it for planning or policy optimization in imagination.
The Dreamer or Recurrent State Space Model (RSSM) architecture~\cite{hafner2019planet,hafner2020dreamer}
has become a standard backbone for world models, enabling both online planning via Model Predictive Control (MPC)~\cite{hafner2019planet} and policy learning with Reinforcement Learning (RL) through imagined rollouts~\cite{hafner2020dreamer,hafner2023dreamerv3}.

A fundamental challenge in model-based RL is
\emph{objective mismatch}~\cite{lambert2020objective}:
the training objective for the world model (prediction accuracy) does not
necessarily align with the downstream objective (closed-loop performance).
We see this directly: on LunarLander, validation loss and multi-step prediction error keep improving while the closed-loop return of CEM-MPC on the same checkpoints may rise, peak and then collapse.

We focus on LunarLander-v3, in which a lander must touch down softly on a pad from a randomized initial state (Appendix~\ref{sec:app-env}). Its $8$-dimensional observation and $4$ discrete actions are small enough to admit direct structural analysis of the learned latent, and its reward is not recoverable from the current observation and action alone, because of potential-based shaping plus terminal events gated by simulator-internal flags (Section~\ref{sec:reward-predictability}).

Our contributions are: (i) a world model trained on human demonstrations for LunarLander, used for CEM-MPC and imagination-trained A2C, and compared with model-free A2C and behaviour cloning (Section~\ref{sec:closed-loop}); (ii) the Reward Observability Fraction (ROF), with the short argument that its observable subspace is exactly the support of the Fisher information that the next observations carry about the latent state (Section~\ref{sec:rof}); (iii) a $z$-only reward-head test: forcing the head onto the posterior-corrected latent raises ROF, and this is the only one of our ROF comparisons that survives a change of latent coordinates (Section~\ref{sec:validity}); and (iv) a negative result: collapse differs between training runs with identical data and configuration, and no offline statistic we tested, ROF included, predicts it (Section~\ref{sec:negative}).

\section{Related Work}
\label{sec:related}

\paragraph{World models and model-based RL.}
The RSSM family (PlaNet~\cite{hafner2019planet}, Dreamer~\cite{hafner2020dreamer} and DreamerV3~\cite{hafner2023dreamerv3}) learns latent dynamics for planning and imagination-based actor--critic training; PETS~\cite{chua2018pets} pairs probabilistic ensemble dynamics with MPC. These methods train by maximizing prediction likelihood and evaluate by downstream task performance.

\paragraph{Objective mismatch and model selection.}
Lambert et al.~\cite{lambert2020objective} showed that optimizing a dynamics model for prediction accuracy does not guarantee good closed-loop performance. Decision-aware objectives such as value-aware model learning~\cite{farahmand2017vaml} change the training loss instead. For choosing among already-trained models, offline model-based RL has relied on validation loss and off-policy evaluation; Yang et al.~\cite{yang2026boms} argue that both are unreliable under distribution shift and select dynamics models with a small online budget instead, in the spirit of active offline policy selection~\cite{konyushova2021active}. Our negative result (Section~\ref{sec:negative}) is consistent with that conclusion, reached independently.

\paragraph{Model exploitation.}
Planners and policies exploit model error: MOPO~\cite{yu2020mopo} and MOReL~\cite{kidambi2020morel} penalize model uncertainty during policy optimization, and Sims et al.~\cite{sims2024edge} show that truncated model rollouts can drive value overestimation and collapse even with an exact dynamics model. We probe several planner-side explanations of this kind in Section~\ref{sec:negative-tests}.

\paragraph{Representations, control theory and Fisher information.}
DeepMDP~\cite{gelada2019deepmdp}, SOLAR~\cite{zhang2019solar} and contrastive behavioural embeddings~\cite{agarwal2021contrastive} shape the representation at training time; we instead analyse a trained model post hoc, with tools that go back to Kalman's controllability and observability~\cite{kalman1963mathematical}, Moore's balanced realizations~\cite{moore1981principal}, and local Jacobian analysis of trained recurrent networks~\cite{sussillo2013opening}. The link between observability and the Fisher information of a linear-Gaussian measurement model is classical~\cite{kay1993fundamentals,anderson1979optimal}; Section~\ref{sec:fisher} states the form we need, which concerns the support of that information rather than its magnitude.

\section{Method}
\label{sec:method}

\subsection{World Model Architecture}
\label{sec:architecture}

We use an RSSM-inspired world model~\cite{hafner2020dreamer,hafner2023dreamerv3} adapted for LunarLander-v3, whose observation is $p{=}8$-dimensional (6 continuous physics dimensions plus 2 binary leg-contact indicators) with $K{=}4$ discrete actions. The RSSM core maintains a deterministic hidden state $\bm{h}_t \in \R^{d_h}$ ($d_h{=}256$, single-layer GRU) and a stochastic latent $\bm{z}_t \in \R^{d_z}$ ($d_z{=}16$), giving a full latent state $\bm{s}_t = [\bm{h}_t;\, \bm{z}_t] \in \R^{272}$. Actions $a_t \in \{0,1,2,3\}$ are one-hot encoded as $\bm{u}_t \in \R^4$. The RSSM transition (prior) is
\begin{equation}
  \bm{s}_{t+1} = f_\theta(\bm{s}_t, \bm{u}_t)
    \;=\;
    \bigl[\operatorname{GRU}(\bm{h}_t,\; [\bm{z}_t;\bm{u}_t]);\;
          \bm{\mu}_{\text{prior}}(\bm{h}_{t+1})\bigr],
  \label{eq:transition}
\end{equation}
the posterior encoder is $q_\theta(\bm{z}_t \mid \bm{h}_t, \bm{o}_t)$, and both prior and posterior are 2-layer MLPs (LayerNorm + SiLU) predicting mean and log-std of a diagonal Gaussian; the observation encoder is a 2-layer MLP mapping $\bm{o}_t \in \R^8$ to a feature vector concatenated with $\bm{h}_t$ for the posterior. Note that the posterior corrects only $\bm{z}_t$: the deterministic path $\bm{h}_t$ receives observations only through the $\bm{z}$ it is fed.

The decoder has a shared MLP backbone feeding physics (MSE), contact (BCE) and done (BCE) heads, $\hat{\bm{o}}_t = g_\theta(\bm{s}_t)$, and a separate 3-layer reward MLP $\hat{r}_t = \rho_\theta(\bm{s}_t)$ that reads $[\bm{h};\bm{z}]$ directly. The model is trained on a weighted sum of reconstruction, reward, done and KL losses (Appendix~\ref{sec:app-model}). All heads are trained on posterior states, so the reward head learns to read the reward of a transition after the posterior has seen the resulting observation.

\subsection{MPC via Cross-Entropy Method (CEM)}
\label{sec:cem}

Since LunarLander-v3 has a discrete action space, we use the Cross-Entropy Method (CEM)~\cite{rubinstein1999cross} adapted for discrete actions: at each real timestep, CEM iteratively refines a per-step categorical distribution over a planning horizon of $H{=}25$ steps. Each iteration samples $N{=}384$ candidate action sequences, rolls each one through the world-model prior open-loop, and scores it by
\begin{equation}
  J(\bm{u}_{t:t+H-1}) \;=\; \sum_{k=1}^{H} \gamma^{k-1}\bigl( w_{\hat r}\,\hat{r}_{t+k} \;-\; w_{\text{done}}\,\hat{p}^{\,\text{done}}_{t+k} \bigr),
  \label{eq:cem-objective}
\end{equation}
with $\gamma{=}0.97$, $w_{\hat r}{=}0.2$ and $w_{\text{done}}{=}2.5$, where $\hat{p}^{\,\text{done}}$ is the done head's predicted termination probability. No hand-designed state cost is used. CEM keeps the top $E{=}48$ elites and updates the per-step logits toward the elite action frequencies with smoothing factor $\alpha{=}0.7$. After $I{=}4$ such iterations, the first action of the best sequence is executed and the procedure repeats at the next real timestep.

\subsection{Actor-Critic Policy via Latent Imagination}
\label{sec:ac}

Following Dreamer~\cite{hafner2020dreamer}, we also train an actor--critic policy (A2C) entirely in the world model's latent imagination: five posterior warm-up steps on logged data, then $15$ imagined prior steps under the actor, with $\lambda$-return targets from the reward head. It uses no real-environment interaction beyond the offline dataset (details in Appendix~\ref{sec:app-model}).

\subsection{Baselines}
\label{sec:mfac-method}
\label{sec:bc-method}

Model-free A2C (MF-AC) uses actor and critic networks of the same width on the raw observation and is trained in the real environment (about $18.4$M transitions). Behaviour cloning (BC) trains the same actor architecture by cross-entropy on the human actions of the $285$ training episodes with return at least $+200$, and is evaluated with both argmax and sampled actions. BC does not use the world model. Details are in Appendix~\ref{sec:app-model}.

\section{The Reward Observability Fraction}
\label{sec:rof}

\subsection{Linearization, observability matrix and Gramian}
\label{sec:jacobian}

We linearize the transition, decoder and reward functions around a latent state $\bm{s} = [\bm{h};\bm{z}] \in \R^n$ ($n{=}272$) with action $\bm{u}$:
\begin{align}
  A &= \frac{\partial f_\theta}{\partial \bm{s}}\biggr|_{\bm{s},\bm{u}}
    \in \R^{n \times n},
  &
  B &= \frac{\partial f_\theta}{\partial \bm{u}}\biggr|_{\bm{s},\bm{u}}
    \in \R^{n \times K},
  \label{eq:AB}
  \\[4pt]
  C &= \frac{\partial g_\theta}{\partial \bm{s}}\biggr|_{\bm{s}}
    \in \R^{p \times n},
  &
  R &= \frac{\partial \rho_\theta}{\partial \bm{s}}\biggr|_{\bm{s}}
    \in \R^{1 \times n}.
  \label{eq:CR}
\end{align}
For a horizon $H$ the $H$-step observability matrix and observability Gramian are
\begin{equation}
  \mathcal{O}_H = \begin{bmatrix} C \\ CA \\ \vdots \\ CA^{H-1} \end{bmatrix} \in \R^{Hp \times n},
  \qquad
  G_H = \mathcal{O}_H^\top \mathcal{O}_H = \sum_{k=0}^{H-1} (A^k)^\top C^\top C A^k .
  \label{eq:gramian}
\end{equation}
The Gramian has a direct interpretation: $\bm{x}^\top G_H \bm{x} = \sum_{k=0}^{H-1} \norm{C A^k \bm{x}}^2$ is the total squared change in the next $H$ predicted observations produced by perturbing the latent state along $\bm{x}$.

The \emph{$H$-step observable subspace} is $\Obs_H = \mathrm{range}(\mathcal{O}_H^\top) = \mathrm{range}(G_H)$, with orthogonal complement $\Obs_H^\perp = \ker \mathcal{O}_H$; write $P_{\Obs}$ for the orthogonal projector onto $\Obs_H$. The controllability matrix $\mathcal{C}_H = [B,\, AB,\, \dots,\, A^{H-1} B]$ gives, analogously, the subspace $\mathrm{range}(\mathcal{C}_H)$ of latent directions that actions can move within $H$ steps. We use $H{=}25$, the CEM planning horizon. All statements are horizon-indexed; no controllability or minimality assumptions are made, and $H$ may be (and here is) much smaller than $n$.

\subsection{Definitions}

With $\bm{r} = R^\top$ the reward gradient, we introduce two novel metrics:
\begin{equation}
  \ROF \;=\; \frac{\norm{P_{\Obs}\, \bm{r}}^2}{\norm{\bm{r}}^2} \;\in\; [0,1],
  \qquad
  \RCF \;=\; \frac{\norm{P_{\mathcal{C}}\, \bm{r}}^2}{\norm{\bm{r}}^2} \;\in\; [0,1],
  \label{eq:rof}
\end{equation}
where $P_{\mathcal{C}}$ projects onto $\mathrm{range}(\mathcal{C}_H)$. ROF (Reward Observability Fraction) is the fraction of the reward gradient's energy in the observable subspace; RCF (Reward Controllability Fraction) is its control-side counterpart, the fraction in the directions actions can move. In this paper, we mostly focus on studying ROF.

\subsection{What ROF measures, intuitively}

Row block $k$ of $\mathcal{O}_H$ answers the question: if the belief were wrong along direction $\bm{x}$, how would the observation $k$ steps later change? A direction that changes \emph{some} observation within the horizon is one that sensing can, in principle, detect; a direction that changes none of them is invisible to the next $H$ observations. $\Obs_H$ is the set of detectable directions. The reward gradient $\bm{r}$ is the direction in latent space that the reward head is most sensitive to.

ROF therefore asks how much of what the reward head reads is made of things the observations reveal. Because it is a ratio of Euclidean lengths, the answer depends on how the latent axes are scaled: stretching one coordinate can move ROF without changing what the model computes. We read it as a property of the trained coordinates, not of the task. The theorem below makes ``reveal'' precise.

\subsection{Observable subspace $=$ Fisher-information support}
\label{sec:fisher}

Consider the linear-Gaussian model obtained from the linearization at a belief state,
\begin{equation}
  \bm{s}_{t+1} = A \bm{s}_t + B \bm{u}_t + \bm{w}_t, \qquad
  \bm{o}_t = C \bm{s}_t + \bm{v}_t, \qquad
  r_t = R\, \bm{s}_t,
  \label{eq:lgss}
\end{equation}
with known inputs $\bm{u}_t$, process noise $\bm{w}_t \sim \mathcal N(0, Q)$, $Q \succeq 0$, and observation noise $\bm{v}_t \sim \mathcal N(0, \Sigma_v)$, $\Sigma_v \succ 0$, independent across time. Stack the observation bundle $\bm{o}_{0:H-1}$, the current observation and the $H{-}1$ that follow. Conditional on the current state $\bm{s}_0$ and the inputs,
\begin{equation}
  \bm{o}_{0:H-1} \;=\; \mathcal{O}_H\, \bm{s}_0 \;+\; \bm{\mu}(\bm{u}_{0:H-2}) \;+\; \bm{\eta},
  \qquad \bm{\eta} \sim \mathcal N(0, \Sigma_o),
  \label{eq:bundle}
\end{equation}
where $\bm{\mu}$ collects the known input contributions and $\bm{\eta}$ the propagated process noise and the observation noise. $\bm{\eta}$ is correlated across time whenever $Q \neq 0$, and $\Sigma_o \succ 0$ regardless of $Q$ because $\Sigma_v \succ 0$. The covariance does not depend on $\bm{s}_0$, so the Fisher information of the bundle about $\bm{s}_0$ is $J = \mathcal{O}_H^\top \Sigma_o^{-1} \mathcal{O}_H$.

\begin{theorem}[Fisher support]\label{thm:support}
For the model~\eqref{eq:lgss} with $\Sigma_v \succ 0$:
\begin{enumerate}
\item[(a)] $\ker J = \ker \mathcal{O}_H$, hence $\mathrm{range}(J) = \Obs_H$, for every admissible noise model $(Q, \Sigma_v)$.
\item[(b)] The likelihood depends on the state only through its observable part:
\[ p(\bm{o}_{0:H-1} \mid \bm{s}_0) = p(\bm{o}_{0:H-1} \mid P_{\Obs}\, \bm{s}_0). \]
Consequently, for any proper prior over $\bm{s}_0$, $I(\bm{s}_0;\, \bm{o}_{0:H-1}) = I(P_{\Obs}\, \bm{s}_0;\, \bm{o}_{0:H-1})$.
\end{enumerate}
\end{theorem}

\begin{proof}
(a) For $\bm{x} \in \R^n$, $\bm{x}^\top J \bm{x} = \norm{\Sigma_o^{-1/2} \mathcal{O}_H \bm{x}}^2$, which vanishes if and only if $\mathcal{O}_H \bm{x} = 0$ because $\Sigma_o^{-1/2}$ is invertible. Since $J \succeq 0$, $\ker J = \{\bm{x} : \bm{x}^\top J \bm{x} = 0\} = \ker \mathcal{O}_H$, and $\mathrm{range}(J) = (\ker J)^\perp = (\ker \mathcal{O}_H)^\perp = \Obs_H$.

(b) By~\eqref{eq:bundle} the conditional law of the bundle given $\bm{s}_0$ is Gaussian with mean $\mathcal{O}_H \bm{s}_0 + \bm{\mu}$ and covariance $\Sigma_o$, which is free of $\bm{s}_0$. Because $\mathcal{O}_H (I - P_\Obs) = 0$, the mean, and hence the whole conditional law, depends on $\bm{s}_0$ only through $P_\Obs \bm{s}_0$. The Markov chain $\bm{s}_0 \to P_\Obs \bm{s}_0 \to \bm{o}_{0:H-1}$ then gives $I(\bm{s}_0; \bm{o}) \le I(P_\Obs \bm{s}_0; \bm{o})$ by data processing, and the reverse inequality holds because $P_\Obs \bm{s}_0$ is a function of $\bm{s}_0$.
\end{proof}

The theorem fixes what the observable subspace means for a trained model, and hence what ROF can and cannot claim.

\begin{remark}[What the theorem does not say]
Part (b) is deliberately prior-free. It does \emph{not} say that the posterior variance of $R\bm{s}$ cannot shrink when $\bm{r} \in \Obs_H^\perp$: a correlated prior lets observable coordinates inform unobservable ones. What holds without qualification is that the likelihood itself is silent about $\Obs_H^\perp$; any learning there is inherited from prior correlation.
\end{remark}

\begin{corollary}[Homoscedastic decoders]\label{cor:mse}
If $\Sigma_v = \sigma^2 I$ and $Q = 0$, then $\Sigma_o = \sigma^2 I$ and $J = \sigma^{-2} G_H$: the unweighted Gramian~\eqref{eq:gramian} recovers not only the support of the Fisher information but its principal directions and, up to $\sigma^{-2}$, its spectrum.
\end{corollary}

An MSE reconstruction loss makes $\Sigma_v = \sigma^2 I$ the model's own assumption. The RSSM's stochastic transition violates $Q = 0$, so for trained models the unweighted construction is guaranteed to recover only the support, by Theorem~\ref{thm:support}(a). The support is all that ROF uses.

\paragraph{Reading.}
ROF is the Euclidean fraction of the reward gradient on that support; $1-\ROF$ is the fraction in directions the observations are silent about. Three cautions. It is not itself a Fisher-information quantity, or a measure of how \emph{much} information the reward direction receives. The Euclidean split is not invariant to rescaling the latent, so raw levels should not be compared across models. And the construction is exact only for the linearization; Section~\ref{sec:nonlinear} checks the trained nonlinear model.

\subsection{Estimation}
\label{sec:estimation}

For each checkpoint we collect latent states by a 5-step posterior warm-up on validation sequences and compute $A$, $B$, $C$, $R$ by automatic differentiation at each state. All Jacobian quantities in this paper, including ROF, use a single fixed linearization per state (powers $A^k$ in~\eqref{eq:gramian}); a time-varying variant that re-linearizes along the rollout is implemented in the released code but is not used in the reported results. The observable basis is the set of right singular vectors of $\mathcal{O}_H$ with singular value above $10^{-3}\sigma_{\max}$; this threshold is the one step with no exact counterpart in the theory. We report two estimates, each averaged over $N_J{=}128$ states: \emph{good-pool} ROF, from validation episodes with return $\ge +100$ ($2{,}713$ sequences), and \emph{bad-pool} ROF, from validation episodes with return $\le -100$ ($1{,}027$ sequences).

Single-checkpoint estimates are noisy (Appendix~\ref{sec:app-repro}), so we report ROF averaged over three metric seeds and smoothed with a centred 7-checkpoint moving average (MA-7).

\section{Closed-Loop Results}
\label{sec:closed-loop}

\subsection{Environment and Data}
\label{sec:data}

We use Gymnasium's LunarLander-v3 (Section~\ref{sec:architecture}). The reward decomposes into per-step shaping plus terminal events; we give the explicit formula and analyse its observability in Section~\ref{sec:reward-predictability}.

We collected $872$ human-piloted episodes ($180{,}916$ environment steps; $158{,}685$ train + $22{,}231$ validation), split $750/122$ train/val ($32{,}038$/$4{,}500$ valid sequence positions at length $30$, stride $5$). Two pilots took turns to introduce stylistic diversity, and the mix deliberately covers clean landings, near-misses, and crashes so the world model sees both reward-positive and reward-negative regimes; the mean return of the training episodes is $-38.2$. The dataset is collected once and reused for all model-based training and evaluation.

\subsection{World Model Training}

The RSSM world model (WM) is trained for $500$ epochs with learning rate $10^{-4}$ (AdamW), batch size $64$, sequence length $30$ at stride $5$, KL free-bits floor $\beta_{\text{KL}}{=}0.5$, and loss weights as in~\eqref{eq:loss}. Checkpoints are saved every $5$ epochs, yielding $100$ for evaluation. Seeds (dataloader shuffling, model initialization, evaluation episodes) are fixed to $12345$ unless stated otherwise.

\paragraph{Runs.} We refer to training runs by tags of the form \emph{seed}-\emph{GPU}-\emph{protocol} (Table~\ref{tab:runs} in Appendix~\ref{sec:app-model}). All downstream experiments in this section use a single run, \texttt{12345-3080-R}, the run analysed in the first version; repeating the A2C and behaviour-cloning comparisons on every run would have been prohibitive. It is not the strongest run: it collapses late in training, and runs with seed $777$ reach higher MPC returns (Figure~\ref{fig:all-runs}).

\subsection{MPC Evaluation}
\label{sec:mpc-perf}

\begin{figure}[htbp]
\centering
\includegraphics[width=0.95\textwidth]{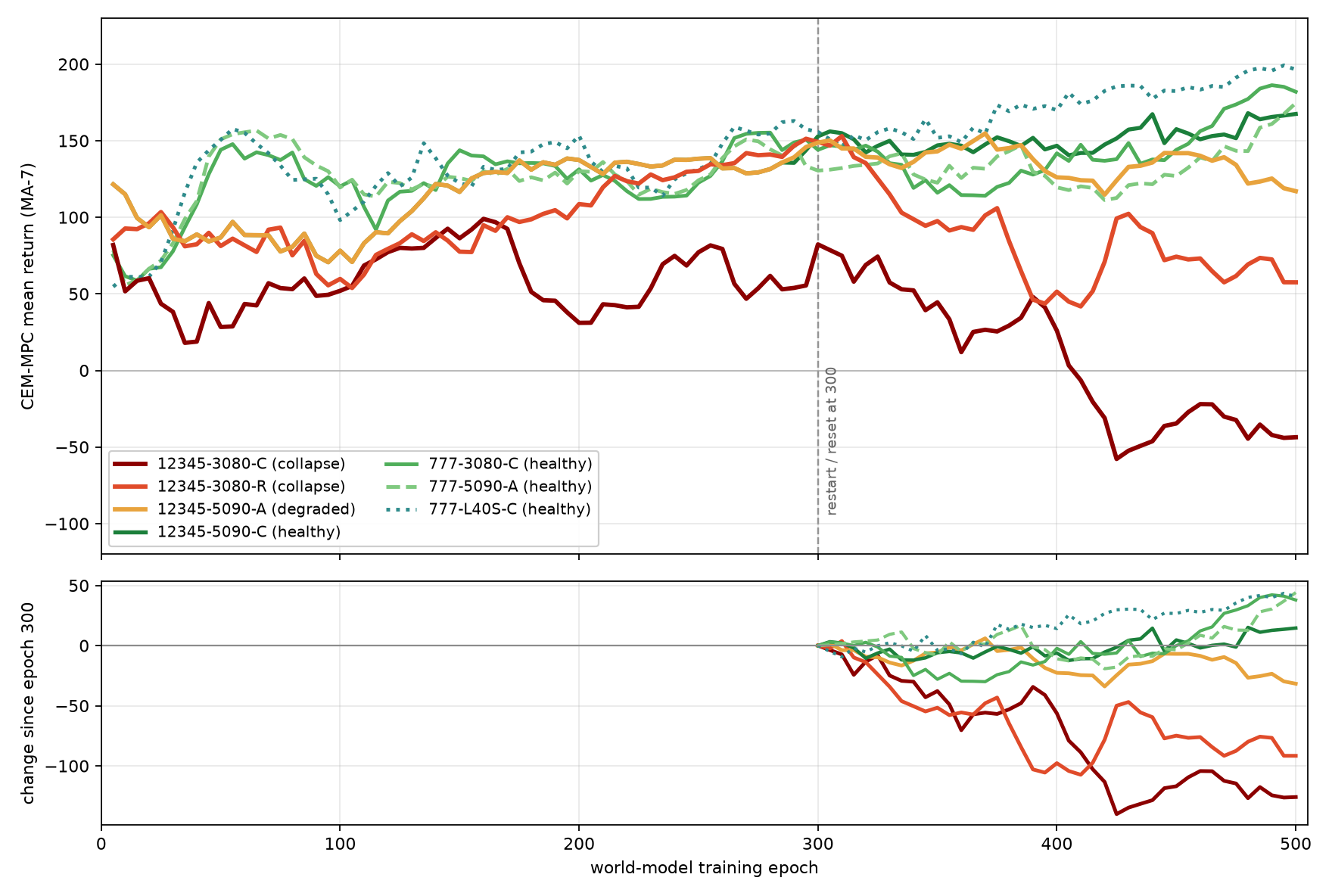}
\caption{MA-7 CEM-MPC return over training for the standard-configuration runs of Table~\ref{tab:runs}. Red: collapsed; amber: degraded; green: healthy. Bottom: change relative to each run's own level at epoch $300$, where \texttt{12345-3080-R} was resumed and the \texttt{-A} runs had their optimizer reset.}
\label{fig:all-runs}
\end{figure}

Each of the $100$ checkpoints is evaluated with CEM-MPC (Section~\ref{sec:cem}) on $20$ episodes (episode $k$ uses environment seed $12345{+}k$, $600$-step cap). Per-checkpoint means are noisy (std $\approx 80$--$120$ across episodes), so we use a centred 7-checkpoint moving average (MA-7) as the reference for closed-loop quality. Figure~\ref{fig:all-runs} shows the resulting curves. The MA-7 return of \texttt{12345-3080-R} peaks at $+153.0$ (epoch $310$) over a plateau at epochs ${\sim}260$--$320$ and collapses after epoch $380$ while validation loss keeps improving; the seed-$777$ runs rise monotonically (Section~\ref{sec:heterogeneity}).

\subsection{Model-Based A2C and Data Efficiency}
\label{sec:ac-setup}
\label{sec:data-efficiency}

We train A2C in imagination (Section~\ref{sec:ac}) on nine checkpoints of \texttt{12345-3080-R}, spanning early training to the MPC plateau, and evaluate every saved actor on $100$ deterministic episodes (seeds $12345{+}k$, $600$-step cap). The plateau checkpoints give the most consistently safe A2C policies; early checkpoints can reach a high best-case mean (world model $100$ has the highest) but their safest policies score much lower (Appendix~\ref{sec:app-ac}, Table~\ref{tab:ac-summary}). Identifying the plateau required the closed-loop MPC sweep. MF-AC is evaluated with the same $100$-episode protocol but LunarLander's standard $1000$-step horizon; on these episodes it peaks at epoch~$760$ with mean $+193.04$ ($54/100$ perfect, $0/100$ catastrophic).

We measure data cost in real environment steps. All model-based methods share the same offline dataset ($180{,}916$ steps); MF-AC's cost is the cumulative transition count up to its best epoch and needs $65.3\times$ more data than world-model based methods. Table~\ref{tab:data-efficiency} reports the results.

\begin{table}[htbp]
\centering
\small
\setlength{\tabcolsep}{4pt}
\resizebox{\textwidth}{!}{%
\begin{tabular}{@{}l r r r r r r@{}}
\toprule
\textbf{Method} & \textbf{Training steps} &  \textbf{Mean} & \textbf{Worst} & \textbf{Perfect} & \textbf{Catast.} & \textbf{Data used} \\
\midrule
MPC (smoothed best, WM~$310$)            & $180{,}916$      & $+166.60$          & $\mathbf{-138.52}$ & $64/100$          & $3/100$           & $\mathbf{1\times}$ \\
WM-AC (WM~$280$, A2C ep~$800$)           & $180{,}916$      & $\mathbf{+189.53}$ & $-419.91$          & $\mathbf{70/100}$ & $5/100$           & $\mathbf{1\times}$ \\
MF-AC (peak @ ep~$760$)                  & $11{,}820{,}726$ & $+183.67$          & $-153.11$          & $49/100$          & $\mathbf{1/100}$  & $65.3\times$ \\
\bottomrule
\end{tabular}%
}
\caption{MPC vs world-model-based AC policy vs model-free AC policy training complexity and closed-loop returns in real environment. Model-free AC policy (MF-AC) is the most expensive to train data-wise (needs $65.3\times$ more data provided as sim interactions). WM-AC matches MF-AC's return despite using only the offline dataset. Checkpoint evaluation of the A2C policies themselves is excluded on both sides. All rows use $100$ deterministic episodes. The two A2C rows are scored on held-out episodes (seeds $99999{+}k$) that were not used to choose their checkpoints; on the selection episodes (seeds $12345{+}k$) they scored $+217.48$ and $+193.04$. MPC uses seeds $12345{+}k$; its checkpoint was chosen from a $20$-episode sweep on the first $20$ of these. MPC and WM-AC use a $600$-step cap; MF-AC uses the standard $1000$-step horizon. \textbf{Perfect} $=$ return $>+200$ (as in the evaluation code); \textbf{Catast.}\ $=$ return $<-100$.}
\label{tab:data-efficiency}
\end{table}

\paragraph{Analysis.}
On $100$ held-out episodes (seeds $99999{+}k$, same caps), the selected model-based A2C policy (WM~$280$) scores $+189.53$ and MF-AC $+183.67$. The difference of $+5.9$ is within noise (paired bootstrap $95\%$ interval $[-23.6, +32.2]$). WM-AC lands perfectly more often ($70/100$ against $49/100$) but has the heavier tail ($5/100$ catastrophic against $1/100$); with a $1000$-step cap for both, WM-AC scores $+196.64$. On the episodes used to select them, the two checkpoints scored $+217.48$ and $+193.04$: each is the best of $20$ A2C checkpoints on those episodes, and selection inflated WM-AC more ($28$ points) than MF-AC ($9$ points). We therefore read WM-AC as matching MF-AC's return with ${\sim}65\times$ fewer real training transitions. CEM-MPC on the same world-model family reaches $+166.6$ with no policy training. At inference time, the A2C policy runs in real-time and CEM-MPC runs at $10$Hz on 5090 GPU.

\subsection{Behaviour Cloning vs.\ World Model with MPC}
\label{sec:bc}

Table~\ref{tab:bc} compares the BC policy of Section~\ref{sec:bc-method} with CEM-MPC. All rows use the same $20$ evaluation episodes (environment seeds $12345{+}k$) and the same $600$-step cap.

\begin{table}[htbp]
\centering
\small
\begin{tabular}{@{}l r r r r r r@{}}
\toprule
\textbf{Policy} & \textbf{Mean} & \textbf{Worst} & \textbf{Perfect} & \textbf{Negative} & \textbf{Catast.} & \textbf{Avg.\ steps} \\
\midrule
BC, deterministic (argmax)             & $-190.2$ & $-681.5$ & $2/20$  & $16$ & $12$ & $363.5$ \\
BC, stochastic (sampled)               & $+185.6$ & $-190.8$ & $15/20$ & $3$  & $1$  & $285.2$ \\
MPC, \texttt{12345-3080-R}, WM~$280$      & $+162.8$ & $-37.5$  & $12/20$ & $4$  & $0$  & $383.2$ \\
MPC, \texttt{12345-3080-R}, WM~$310$ & $+156.0$ & $-70.9$ & $12/20$ & $4$ & $0$  & $380.0$ \\
MPC, best checkpoint of all runs$^\dagger$ & $+216.6$ & $+12.7$ & $18/20$ & $0$ & $0$  & $312.4$ \\
\bottomrule
\end{tabular}
\caption{Behaviour cloning vs.\ CEM-MPC, $20$ episodes each on identical initial conditions. \textbf{Perfect}: return $>+200$; \textbf{Negative}: return $<0$; \textbf{Catast.}: return $<-100$. $^\dagger$\texttt{777-5090-A}, epoch $500$ (Table~\ref{tab:runs}); this is the best of $100$ checkpoints on noisy $20$-episode means and is therefore selection-biased upward. The BC checkpoint was chosen by validation loss with no environment interaction.}
\label{tab:bc}
\end{table}

Three observations. First, \textbf{action selection decides whether BC works at all}: the same weights score $+185.6$ when sampling and $-190.2$ under argmax. The human pilots control the main engine by duty cycle, firing it on $42.3\%$ of steps in the successful demonstrations; the policy learns that firing \emph{rate} as a probability, and argmax turns a probability such as $0.55$ into firing on every step, a different controller that over-thrusts and drifts off the demonstrated state distribution. Second, \textbf{stochastic BC matches MPC on the mean, and on these $20$ episodes its tail is worse}: every MPC row has zero catastrophic episodes and a worst case of $-70.9$ or better, while BC has one catastrophic episode and a worst case of $-190.8$. BC also descends visibly faster and more roughly ($285$ steps per episode against $380$--$383$ for MPC on \texttt{12345-3080-R}). Third, BC is only trained on good successful landing episodes and therefore its training data is harder to obtain compared to world-model based methods.

\section{What Moves ROF}
\label{sec:validity}

The theorem names a subspace. ROF is a Euclidean split of the reward gradient onto that subspace, so it depends on how the latent is scaled. To separate geometry from scaling we recompute ROF after whitening the latent by its covariance over validation states, which makes the fraction invariant to any invertible linear change of latent coordinates (Appendix~\ref{sec:app-coords}). Three comparisons drawn from raw ROF do not survive this (Table~\ref{tab:coords}). Across twelve runs (four training-data mixtures and two capacity variants, two seeds each), the rank correlation between raw and whitened bad-pool levels is $-0.09$. Halving the recurrent state shifts the raw level by $+0.098$ but the whitened level by only $+0.007$. The within-run correlation between ROF and MPC return, positive in all eight Reacher runs and of mixed sign on LunarLander in raw coordinates, changes under whitening to negative in all eight Reacher runs and positive in all eight LunarLander runs. We therefore do not use capacity or domain comparisons of ROF as evidence. The test that remains is a hard restriction of the reward head.

\subsection{Restricting the reward head to the stochastic latent}
\label{sec:zonly}

The posterior corrects only $\bm{z}$ (Section~\ref{sec:architecture}). We retrain the world model with the reward head reading $\bm{z}$ alone: its first layer becomes $\mathbb{R}^{16}{\to}\mathbb{R}^{256}$ instead of $\mathbb{R}^{272}{\to}\mathbb{R}^{256}$, a genuine truncation of its input rather than a zero mask, swapped in after the rest of the network is built so that every other module receives the same initialization as the ordinary run of that seed. Everything else is unchanged (seed $12345$, one continuous 500-epoch process, no optimizer reset). We run this ablation on two GPUs (\texttt{12345-5090-Z} and \texttt{12345-3080-Z}), each paired with the ordinary $[\bm{h},\bm{z}]$ head on the same seed and GPU (\texttt{12345-5090-C} and \texttt{12345-3080-C}).

\begin{table}[htbp]
\centering
\small
\setlength{\tabcolsep}{4pt}
\begin{tabular}{@{}l l r r r r r r@{}}
\toprule
& & \multicolumn{3}{c}{\textbf{CEM-MPC return}} & \multicolumn{3}{c}{\textbf{Geometry (epochs $\ge 400$)}} \\
\cmidrule(lr){3-5}\cmidrule(lr){6-8}
\textbf{Run} & \textbf{Reward head} & \textbf{MA-7 peak} & \textbf{Final-10} & \textbf{Late trend} & \textbf{ROF} & \textbf{RCF} & $\bm{\rho_s}$\textbf{(ROF, MPC)} \\
\midrule
\texttt{12345-5090-C} & $[\bm{h},\bm{z}]$ & $+168.0$ @ 480 & $+157.6$ & $+0.57$ & $0.221$ & $0.261$ & $-0.23$ \\
\texttt{12345-5090-Z} & $\bm{z}$ only     & $+143.0$ @ 305 & $+124.1$ & $+0.29$ & $\mathbf{0.280}$ & $\mathbf{0.184}$ & $\mathbf{+0.80}$ \\
\midrule
\texttt{12345-3080-C} & $[\bm{h},\bm{z}]$ & $+99.0$ @ 160  & $-31.9$  & $-0.88$ & $0.196$ & $0.269$ & $+0.27$ \\
\texttt{12345-3080-Z} & $\bm{z}$ only     & $+107.7$ @ 240 & $+15.1$  & $-0.87$ & $\mathbf{0.284}$ & $\mathbf{0.189}$ & $+0.26$ \\
\bottomrule
\end{tabular}
\caption{Reward-head ablation, seed $12345$. MPC: $20$ episodes per checkpoint; peak on the MA-7 curve (epochs $>50$), final-10 as the mean of the last ten raw checkpoint means, late trend as Spearman $\rho$ between epoch and MA-7 return over epochs $>300$. Geometry: good-pool ROF and RCF averaged over three metric seeds and over epochs $\ge 400$; last column is the within-run Spearman correlation between MA-7 good-pool ROF and MA-7 MPC return (epochs $>50$).}
\label{tab:zonly}
\end{table}

\begin{figure}[htbp]
\centering
\includegraphics[width=0.78\textwidth]{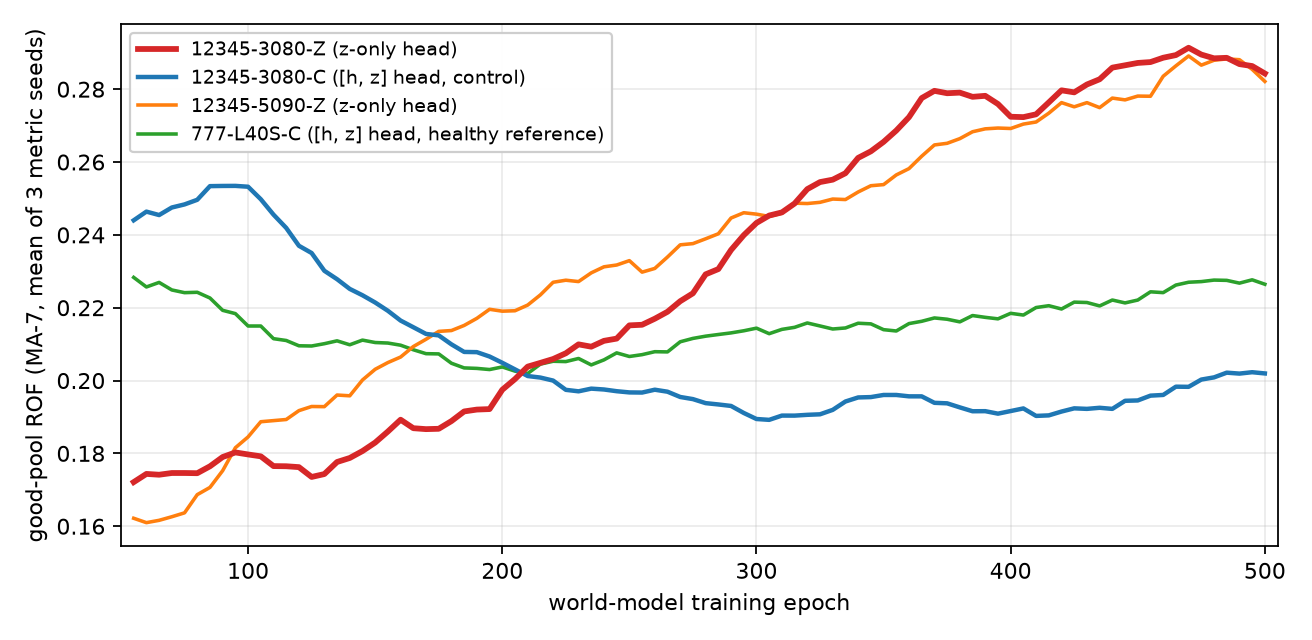}
\caption{Good-pool ROF over training. Restricting the reward head to $\bm{z}$ raises ROF on both GPUs, to almost the same level, while the $[\bm{h},\bm{z}]$ control \texttt{12345-3080-C} falls. \texttt{12345-3080-Z} climbs throughout the epochs in which its closed-loop return collapses (Figure~\ref{fig:zonly-mpc}).}
\label{fig:zonly-rof}
\end{figure}

Table~\ref{tab:zonly} and Figure~\ref{fig:zonly-rof} show the same move on both GPUs: the head is forced onto $\bm{z}$, and more of the reward gradient sits in the observable subspace (and less in the controllable one). On the 3080 GPU pair, whitening the latent (Appendix~\ref{sec:app-coords}) keeps the sign of that gap and roughly halves it. The ablation also costs return: \texttt{12345-5090-Z} is slow to take off and never reaches the ceiling of its $[\bm{h},\bm{z}]$ control, which is consistent with part of LunarLander's reward being read from history in $\bm{h}$, though the ablation does not show what the head reads there.

The 3080 pair was a mechanism test for collapse; it came back negative (Section~\ref{sec:negative}).

\subsection{Reward structure, not ROF}
\label{sec:reward-predictability}
\label{sec:domain}

LunarLander's per-step reward, taken from the Gymnasium reference implementation,\footnote{\url{https://github.com/Farama-Foundation/Gymnasium/blob/main/gymnasium/envs/box2d/lunar_lander.py}} is, on non-terminal steps,
\begin{equation}
r_t \;=\; \underbrace{\phi(o_{t+1}) - \phi(o_t)}_{\text{PBRS shaping}} \;-\; c_{\text{main}}\, \mathbb{1}[a_t{=}\text{main}] \;-\; c_{\text{side}}\, \mathbb{1}[a_t{\in}\{\text{L,R}\}],
\label{eq:lander-reward}
\end{equation}
with shaping potential $\phi(o) = -100(\|p\|+\|v\|) - 100|\theta| + 10(\text{leg}_1+\text{leg}_2)$ defined on the visible state and $(c_{\text{main}}, c_{\text{side}}) = (0.30, 0.03)$. On the terminal step this reward is replaced, not supplemented: by $-100$ after a crash or when the lander leaves the screen ($|x| \ge 1$), and by $+100$ when it comes to rest. Leaving the screen is visible in $o_t$; the crash and rest conditions are Box2D-internal flags (\texttt{game\_over}, \texttt{lander.awake}) that are not. MLP regressors trained on our dataset explain only $R^2{=}0.29$ of the per-step reward from $(o_t, a_t)$, but $0.97$ once $o_{t+1}$ is added and terminal steps are removed. On Reacher with the same RSSM, $(o_t, a_t)$ alone gives $R^2{=}0.9998$.

What remains of the domain contrast does not use ROF. On the same eight seeds, five of eight LunarLander runs collapse and none of the Reacher runs do. LunarLander's reward is only partly in the current observation; Reacher's is almost entirely there. That is a fact about the tasks, not about a Euclidean split in latent space.

\subsection{Validity at the nonlinear model}
\label{sec:nonlinear}

Theorem~\ref{thm:support} is exact only for the linearization. On five checkpoints of \texttt{12345-3080-R}, with twelve belief states each and an acceptance bound of $0.15$ on the response ratio, we perturbed belief states by $\varepsilon \in \{0.05, 0.1, 0.3\}$ along deep-kernel directions (well inside $\ker \mathcal{O}_H$) and along observable directions of matched magnitude, then rolled the model forward open-loop for $H{=}25$ steps. The median ratio of observation-space responses (kernel over observable) was $0.012$--$0.019$ at every $\varepsilon$ and every checkpoint, an order of magnitude inside that bound and flat in $\varepsilon$. Within the states and perturbation sizes we tested, the trained nonlinear model is close to silent about the kernel as well, not only its linearization. A further probe, of how the amortized posterior corrects errors, is in Appendix~\ref{sec:app-probes}.

\section{Offline Prediction of Closed-Loop Collapse: A Negative Result}
\label{sec:negative}

The first version of this paper reported that ROF, computed on a single run (\texttt{12345-3080-R}), tracked the rise--peak--collapse of MPC return ($\rho_s{=}-0.71$) and could select a good checkpoint offline. This section reports what happened when that claim was tested on more LunarLander runs: it does not hold, and neither does any alternative metric we tried.

\subsection{Collapse differs between runs}
\label{sec:heterogeneity}

Three facts about collapse frame the negative result.

\textbf{Collapse is common.} In an eight-seed panel trained with identical code, five of eight runs collapse; in a twelve-run panel, three of the six all-human runs do. On the same eight seeds, Reacher never collapses.

\textbf{On-data metrics do not see it.} Validation loss keeps improving through collapse (Figure~\ref{fig:all-runs}), and so does open-loop cumulative-reward error in $10$ of the $12$ panel runs; the two exceptions are the scripted-data runs, which were never good (Table~\ref{tab:negative}). Among the runs of Table~\ref{tab:zonly} and \texttt{777-L40S-C}, the one with the \emph{worst} late validation loss ($35.98$, \texttt{777-L40S-C}) has by far the best return ($+189.1$ final-10), and \texttt{12345-5090-Z} has lower late validation loss than its control \texttt{12345-5090-C} ($32.75$ vs.\ $34.68$) despite its lower return.

\textbf{Collapse differs between runs with the same configuration.} The seed-$777$ runs rise monotonically on the same data and configuration as \texttt{12345-3080-R} (Figure~\ref{fig:all-runs}). Seed $12345$ trained as one continuous process collapses on an RTX 3080 (\texttt{12345-3080-C}: MA-7 peak $+99.0$, final-10 $-31.9$) but not on an RTX 5090 (\texttt{12345-5090-C}: peak $+168.0$, final-10 $+157.6$). Resetting the AdamW moments at epoch $300$ on a healthy path degrades seed $12345$ only moderately (\texttt{12345-5090-A}) and leaves seed $777$ healthy (\texttt{777-5090-A}), so the restart of \texttt{12345-3080-R} was not the cause. Collapse therefore differs between runs with the same data and nominal configuration, and these comparisons do not identify which difference in the training path causes it. That is why the first version's correlation, measured on one run, did not generalize.

\subsection{ROF does not predict collapse}
\label{sec:rof-negative}

\begin{table}[htbp]
\centering
\small
\begin{tabular}{@{}l l r@{}}
\toprule
\textbf{Run} & \textbf{Outcome} & $\bm{\rho_s}$\textbf{(ROF, MPC)} \\
\midrule
\texttt{12345-3080-R} & collapse & $-0.71$ \\
\texttt{12345-5090-A} & degraded & $-0.50$ \\
\texttt{12345-5090-C} & healthy & $-0.26$ \\
\texttt{12345-3080-C} & collapse & $+0.08$ \\
\texttt{777-5090-A} & healthy & $+0.12$ \\
\texttt{777-3080-C} & healthy & $+0.19$ \\
\texttt{777-L40S-C} & healthy & $+0.76$ \\
\texttt{12345-3080-Z} & collapse & $+0.32$ \\
\texttt{12345-5090-Z} & healthy & $+0.79$ \\
Eight-seed CPU panel & 5 of 8 collapse & $-0.55$ to $+0.32$ \\
\bottomrule
\end{tabular}
\caption{Within-run Spearman correlation between pool-averaged ROF in the trained coordinates and MA-7 MPC return. Table~\ref{tab:zonly} uses good-pool ROF for its four runs, so its values differ. Jacobian-state count and metric seeds differ between runs, so the values are indicative; the spread of signs is not.}
\label{tab:sign}
\end{table}

\paragraph{No stable sign.} Table~\ref{tab:sign} shows that the within-run correlation between ROF and closed-loop return ranges from $-0.71$ to $+0.79$ and crosses zero among both collapsing and healthy runs. The first version's $-0.71$ is one draw from this distribution, and no fixed signed threshold can mean the same thing on two runs. This instability belongs to raw ROF. After whitening (Appendix~\ref{sec:app-coords}), the within-run correlation is positive in $20$ of the $22$ LunarLander runs we recomputed, the other two at $-0.06$. That observation was made after the fact, and we have not tested whitened ROF as a predictor of collapse.

\paragraph{No early warning.} Where ROF has an interior turning point it follows the MPC turn rather than leading it (by $90$--$155$ epochs on the runs where we can check); the lead of ${\sim}60$ epochs on \texttt{12345-3080-R} is the only case in which ROF turned first. In three of the five collapsing runs of the eight-seed panel, ROF's minimum falls at the start of training (epochs $5$ to $30$), so there is no interior turn to give warning at all.

\paragraph{A fixed detector failed out of sample.} We fixed a detector on the $19$ runs available at the time and applied it unchanged to later runs: flag a run if the maximum drawdown of its smoothed good-pool ROF from its trailing maximum exceeds $0.05$ (\texttt{12345-3080-R} and \texttt{12345-3080-C} scored $0.061$ and $0.064$, healthy runs at most $0.044$). On a later twelve-run panel it produced two true positives, one false alarm on a never-good run, and six misses, including a run whose MA-7 return fell from $+151.8$ to a final-10 of $+8.8$ at a drawdown of only $0.035$. On \texttt{12345-3080-Z}, which fell from $+107.7$ to a final-10 of $+15.1$, it stayed quiet at $0.007$.

\begin{figure}[htbp]
\centering
\includegraphics[width=0.78\textwidth]{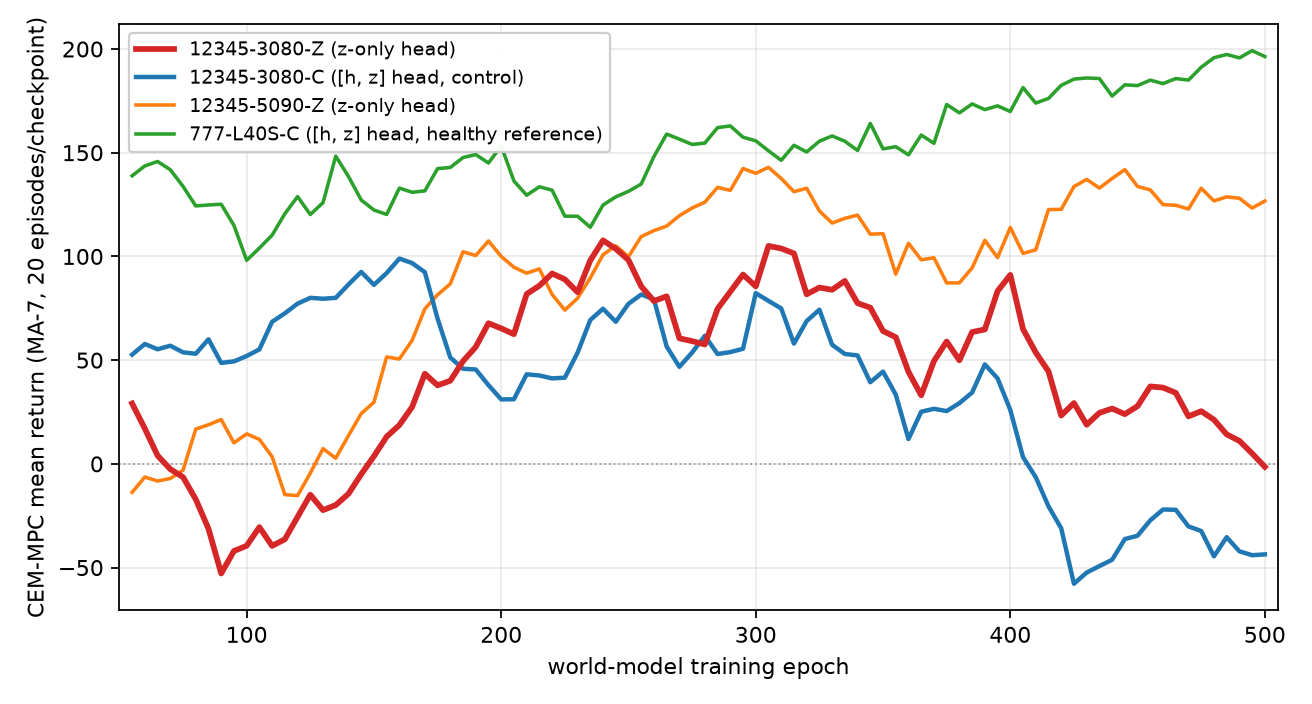}
\caption{MA-7 CEM-MPC return for the reward-head ablation. \texttt{12345-3080-Z} collapses with the same late slope as its $[\bm{h},\bm{z}]$ control \texttt{12345-3080-C}, while its raw ROF rises (Figure~\ref{fig:zonly-rof}).}
\label{fig:zonly-mpc}
\end{figure}

\paragraph{The clearest dissociation.} \texttt{12345-3080-Z} asks whether collapse is caused by reward information sitting in $\bm{h}$. Removing $\bm{h}$ from the reward head does not stop it. Its late trend (the Spearman correlation of MA-7 return with epoch) matches the control's (Table~\ref{tab:zonly}); the peak is later, the decay is the same (Figure~\ref{fig:zonly-mpc}). Raw ROF rises through the collapse (Figure~\ref{fig:zonly-rof}), but whitened ROF stays flat, so that half of the dissociation holds only in the trained coordinates.

\subsection{None of the other offline statistics we tested predicted it}
\label{sec:negative-tests}

Table~\ref{tab:negative} summarizes every test.

\begin{table*}[htbp]
\centering
\footnotesize
\setlength{\tabcolsep}{4pt}
\renewcommand{\arraystretch}{1.15}
\begin{tabular}{@{}p{3.5cm} p{5.1cm} p{1.7cm} p{5.3cm}@{}}
\toprule
\textbf{Test} & \textbf{What was measured} & \textbf{Runs} & \textbf{Result} \\
\midrule
ROF drawdown detector & Max drawdown of smoothed good-pool ROF $> 0.05$ & 12 + 1 held out & 2 true positives, 1 false alarm, 7 missed collapses \\
Offline metric screen & 27 offline metrics (Appendix~\ref{sec:app-metrics}) $\times$ 4 summaries (mean, final, drawdown, trend) $=108$ candidates, against MPC drawdown and final-10 & 17 (11 human-data) & $0$ of $108$ beat a permutation null on the maximum $|\rho_s|$ (thresholds $0.755$--$0.882$); best candidate \texttt{ol\_cumrew\_err.trend}, $|\rho_s|{=}0.82$ \\
Held-out test of best candidate & Trend of open-loop cumulative-reward error & 12 new & $\rho_s{=}-0.17$ and $-0.09$; the error \emph{improves} through collapse in 10 of 12 runs \\
Reward-head mechanism & Remove $\bm{h}$ from the reward head on a collapsing substrate & 2 & Late trend $-0.87$ vs.\ $-0.88$; collapse not prevented \\
Planner leaves the data support & From $59$ pre-contact validation states, CEM plan rolled out in imagination and in the real environment; standardized distance of imagined observations from training observations & 9 & Support distance flat in every run ($|\Delta| \le 0.015$ on $0.37$); optimism rises in 2 of 3 matched pairs; no within-run tracking \\
Planner advantage & Predicted return of the CEM plan minus predicted return of the logged human actions, same states & 9 & $59$--$64$ in every run (means over epochs $100$--$200$ and $400$--$500$; single checkpoints $56$--$68$), healthy or collapsing \\
Closed-loop belief & Instrumented MPC episodes: planner's predicted vs.\ realized return and belief error, flight vs.\ near-ground & 9 & Near-ground belief error rises in every run, healthy ones included; 0 of 3 matched pairs show more optimism in the collapsing arm \\
Terminal-reward fidelity & Imagined reward and termination at logged crashes and landings, $1$--$25$ steps ahead & 8 & Crash anticipated at $-7$ to $-10$ (true $-100$) in every run; no change tracks MPC drawdown ($|\rho_s| \le 0.38$) \\
\bottomrule
\end{tabular}
\caption{Tests of offline and planner-side predictors of closed-loop collapse. ``Runs'' counts LunarLander training runs; each test uses all checkpoints of each run or a fixed stride through them.}
\label{tab:negative}
\end{table*}

Two points deserve emphasis. First, the metrics that evaluate the model on the data distribution improve through collapse: the open-loop cumulative-reward error keeps falling while planning with the same model fails. Second, the probes that look at the planner (where its plans go, how optimistic it is, what it believes near the ground) do not separate healthy from collapsing runs either, with two limits: the support probe starts before leg contact and cannot see the landing, and the closed-loop probe uses ten episodes per checkpoint, which leaves wide scatter. 

\section{Discussion and Limitations}
\label{sec:discussion}

\paragraph{Why offline metrics may be structurally blind here.} Every offline statistic we computed evaluates the model on states drawn from the logged data, or from short rollouts started at them. Collapse, by contrast, is a property of hundreds of closed-loop steps under the planner's own actions, and on this task it varies with differences in the training path that leave on-data prediction quality intact or even improving. A metric that is a function of the model and the validation data alone can only predict it if those differences leave a trace on the data distribution, and in our runs we found none. What makes a model good for planning also depends on where the task's return is concentrated. Here it is concentrated in rare terminal events that imagination barely reproduces: the reward head predicts about $-80$ at a logged crash when the posterior sees the crash, but only $-7$ to $-10$ when imagining it $25$ steps ahead, in every run, so CEM steers on the dense shaping reward, which favours the slow, straight descents we observe. Only the most severe collapse (\texttt{12345-3080-C}) showed a closed-loop signature, a planner whose predicted near-ground return doubled while its realized return fell to $-17$. Both observations argue that offline checkpoint diagnostics are unlikely to be task-agnostic, and that a small budget of closed-loop evaluation, as in~\cite{yang2026boms}, is the practical answer.

\paragraph{What ROF is for.} The theorem names the support (Theorem~\ref{thm:support}). ROF is a Euclidean fraction on that support in the trained coordinates. The $z$-only head moves it, and on the pair we whitened the move survives a change of coordinates; comparisons across capacity and domain do not, and it does not predict collapse. A natural next test is one in which observability actually binds (pixels, occlusion or sensor noise) rather than LunarLander, whose observation is the full physical state. A further test would measure filtering error along reward-relevant directions.

\paragraph{Limitations.} All closed-loop results are on LunarLander-v3, with Reacher as a single contrasting task. The collapsing runs are few and several share seed $12345$; the offline screen has $17$ runs and the held-out test $12$, and its $108$ candidates are four summaries of $27$ related quantities, far from independent. MPC returns use $20$ episodes per checkpoint, the planner probes before leg contact cannot see the landing phase, and the behaviour-cloning comparison uses $20$ episodes against MPC checkpoints chosen by closed-loop evaluation. Because the same seed and configuration can collapse on one GPU and not on another, a single training run should not be read as representative (reproducibility details in Appendix~\ref{sec:app-repro}).

\section{Conclusion}
\label{sec:conclusion}

On LunarLander-v3, an RSSM world model trained on human demonstrations supports a model-based A2C policy that matches the best model-free checkpoint on held-out episodes with far fewer training transitions, and a zero-shot CEM planner that had no catastrophic episodes where behaviour cloning had one, on the same $20$ episodes. The Reward Observability Fraction is the Euclidean share of the reward readout on the Fisher support of the observations, measured in the trained coordinates. Restricting the reward head to $\bm{z}$ moves it. It does not predict when MPC will collapse, and none of the other offline summaries we tested yielded a validated predictor. On this task, the planner has to be tested in closed loop.

\section*{Acknowledgments}
We thank Larry Jackel, Urs Muller, Alexander Popov, and Alexey Kamenev for their careful reviews of the first version. Computational resources were provided by the University of Cape Town's ICTS High Performance Computing facility (\url{hpc.uct.ac.za}).

\bibliographystyle{plain}
\bibliography{references}

\begin{thebibliography}{10}

\bibitem{agarwal2021contrastive}
Rishabh Agarwal, Marlos~C. Machado, Pablo~Samuel Castro, and Marc~G. Bellemare.
\newblock Contrastive behavioral similarity embeddings for generalization in reinforcement learning.
\newblock In {\em International Conference on Learning Representations (ICLR)}, 2021.

\bibitem{anderson1979optimal}
Brian D.~O. Anderson and John~B. Moore.
\newblock {\em Optimal Filtering}.
\newblock Prentice-Hall, 1979.

\bibitem{chua2018pets}
Kurtland Chua, Roberto Calandra, Rowan McAllister, and Sergey Levine.
\newblock Deep reinforcement learning in a handful of trials using probabilistic dynamics models.
\newblock In {\em Advances in Neural Information Processing Systems (NeurIPS)}, volume~31, 2018.

\bibitem{farahmand2017vaml}
Amir-massoud Farahmand, Andre Barreto, and Daniel Nikovski.
\newblock Value-aware loss function for model-based reinforcement learning.
\newblock In {\em Proceedings of the 20th International Conference on Artificial Intelligence and Statistics (AISTATS)}, 2017.

\bibitem{gelada2019deepmdp}
Carles Gelada, Saurabh Kumar, Jacob Buckman, Ofir Nachum, and Marc~G. Bellemare.
\newblock {DeepMDP}: Learning continuous latent space models for representation learning.
\newblock In {\em International Conference on Machine Learning (ICML)}, pages 2170--2179. PMLR, 2019.

\bibitem{hafner2020dreamer}
Danijar Hafner, Timothy Lillicrap, Jimmy Ba, and Mohammad Norouzi.
\newblock Dream to control: Learning behaviors by latent imagination.
\newblock In {\em International Conference on Learning Representations (ICLR)}, 2020.

\bibitem{hafner2019planet}
Danijar Hafner, Timothy Lillicrap, Ian Fischer, Ruben Villegas, David Ha, Honglak Lee, and James Davidson.
\newblock Learning latent dynamics for planning from pixels.
\newblock In {\em International Conference on Machine Learning (ICML)}, pages 2555--2565. PMLR, 2019.

\bibitem{hafner2023dreamerv3}
Danijar Hafner, Jurgis Pasukonis, Jimmy Ba, and Timothy Lillicrap.
\newblock Mastering diverse domains through world models.
\newblock {\em arXiv preprint arXiv:2301.04104}, 2023.

\bibitem{kalman1963mathematical}
Rudolf~E. Kalman.
\newblock Mathematical description of linear dynamical systems.
\newblock {\em Journal of the Society for Industrial and Applied Mathematics, Series A: Control}, 1(2):152--192, 1963.

\bibitem{kay1993fundamentals}
Steven~M. Kay.
\newblock {\em Fundamentals of Statistical Signal Processing, Volume I: Estimation Theory}.
\newblock Prentice Hall, 1993.

\bibitem{kidambi2020morel}
Rahul Kidambi, Aravind Rajeswaran, Praneeth Netrapalli, and Thorsten Joachims.
\newblock {MOReL}: Model-based offline reinforcement learning.
\newblock In {\em Advances in Neural Information Processing Systems}, volume~33, 2020.

\bibitem{konyushova2021active}
Ksenia Konyushova, Yutian Chen, Thomas Paine, Caglar Gulcehre, Cosmin Paduraru, Daniel~J. Mankowitz, Misha Denil, and Nando de~Freitas.
\newblock Active offline policy selection.
\newblock In {\em Advances in Neural Information Processing Systems}, volume~34, 2021.

\bibitem{lambert2020objective}
Nathan Lambert, Brandon Amos, Omry Yadan, and Roberto Calandra.
\newblock Objective mismatch in model-based reinforcement learning.
\newblock In {\em Proceedings of the 2nd Conference on Learning for Dynamics and Control (L4DC)}, pages 761--770. PMLR, 2020.

\bibitem{moore1981principal}
Bruce~C. Moore.
\newblock Principal component analysis in linear systems: Controllability, observability, and model reduction.
\newblock {\em IEEE Transactions on Automatic Control}, 26(1):17--32, 1981.

\bibitem{rubinstein1999cross}
Reuven~Y. Rubinstein.
\newblock The cross-entropy method for combinatorial and continuous optimization.
\newblock {\em Methodology and Computing in Applied Probability}, 1(2):127--190, 1999.

\bibitem{schulman2015gae}
John Schulman, Philipp Moritz, Sergey Levine, Michael~I. Jordan, and Pieter Abbeel.
\newblock High-dimensional continuous control using generalized advantage estimation.
\newblock {\em arXiv preprint arXiv:1506.02438}, 2015.

\bibitem{sims2024edge}
Anya Sims, Cong Lu, Jakob~N. Foerster, and Yee~Whye Teh.
\newblock The edge-of-reach problem in offline model-based reinforcement learning.
\newblock In {\em Advances in Neural Information Processing Systems}, volume~37, 2024.

\bibitem{sussillo2013opening}
David Sussillo and Omri Barak.
\newblock Opening the black box: Low-dimensional dynamics in high-dimensional recurrent neural networks.
\newblock {\em Neural Computation}, 25(3):626--649, 2013.

\bibitem{yang2026boms}
Yuwei Yang, Wei Hung, Yun-Ming Chan, Xi~Liu, and Ping-Chun Hsieh.
\newblock Enhancing offline model-based {RL} via active model selection: A {B}ayesian optimization perspective.
\newblock {\em Machine Learning}, 2026.
\newblock arXiv:2502.11480.

\bibitem{yu2020mopo}
Tianhe Yu, Garrett Thomas, Lantao Yu, Stefano Ermon, James Zou, Sergey Levine, Chelsea Finn, and Tengyu Ma.
\newblock {MOPO}: Model-based offline policy optimization.
\newblock In {\em Advances in Neural Information Processing Systems}, volume~33, 2020.

\bibitem{zhang2019solar}
Marvin Zhang, Sharad Vikram, Laura Smith, Pieter Abbeel, Matthew~J. Johnson, and Sergey Levine.
\newblock {SOLAR}: Deep structured representations for model-based reinforcement learning.
\newblock In {\em International Conference on Machine Learning (ICML)}, pages 7444--7453. PMLR, 2019.

\end{thebibliography}

\clearpage
\appendix
\section{LunarLander-v3 Environment}
\label{sec:app-env}

\begin{figure}[htbp]
\centering
\includegraphics[width=0.55\textwidth]{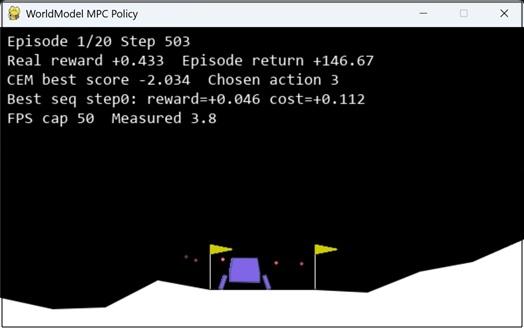}
\caption{LunarLander-v3: four discrete actions (NOP, left/main/right thruster), $8$-D observation, soft-landing objective from a randomized initial state.}
\label{fig:lander}
\end{figure}

In Gymnasium's LunarLander-v3 the agent fires one of four discrete engines (none, left, main, right) to bring a lander from a randomized initial state to a soft touchdown on the pad between the flags, observing an $8$-dimensional state (Figure~\ref{fig:lander}).

\section{Model and Training Details}
\label{sec:app-model}

\paragraph{Decoder.}
Rather than treating all 8 observation dimensions identically, the decoder uses a shared 2-layer MLP backbone (LayerNorm + SiLU, input $d_h{+}d_z{=}272$) feeding four heads: an MSE-regression \emph{physics} head ($\R^6$, continuous states), a BCE \emph{contact} head ($\R^2$ logits, sigmoid at inference), a BCE \emph{done} head ($\R^1$), and a separate 3-layer \emph{reward} MLP ($\R^{272}{\to}256{\to}256{\to}1$) that operates directly on $[\bm{h};\bm{z}]$ rather than the shared backbone, so the reward predictor develops its own representation without competing with reconstruction for backbone capacity. The observation decoder is $\hat{\bm{o}}_t = g_\theta(\bm{s}_t)$ (physics concatenated with sigmoid of contact logits); the reward predictor is $\hat{r}_t = \rho_\theta(\bm{s}_t)$.

\paragraph{Training loss.} The total loss is a weighted sum:
\begin{equation}
  \mathcal{L} = w_r \mathcal{L}_{\text{recon}} + w_\rho \mathcal{L}_{\text{rew}}
    + w_d \mathcal{L}_{\text{done}} + w_{\text{KL}} \, \mathcal{L}_{\text{KL}},
  \label{eq:loss}
\end{equation}
with weights $w_r{=}1.0$, $w_\rho{=}1.2$, $w_d{=}0.5$, $w_{\text{KL}}{=}1.0$
(KL free-bits floor $\beta_{\text{KL}}{=}0.5$).
$\mathcal{L}_{\text{recon}}$ combines MSE for physics and BCE for contacts;
$\mathcal{L}_{\text{rew}}$ is squared reward error on raw (unscaled) rewards;
$\mathcal{L}_{\text{done}}$ is done-flag BCE;
$\mathcal{L}_{\text{KL}} = \text{KL}(q_\theta \| p_\theta)$
is the posterior--prior KL divergence.

\paragraph{Actor-critic in imagination.} Following Dreamer~\cite{hafner2020dreamer}, we train an actor--critic policy (A2C) entirely within the latent imagination of the world model. The actor and critic are compact 3-layer MLPs (hidden dimension of 256) that read the latent state $[\bm{h};\bm{z}]$; the actor outputs a categorical distribution over the 4 actions, the critic predicts scalar value estimates. Each training step samples a real observation--action sequence from the replay dataset, warms the RSSM posterior up for 5 steps on the real data, then imagines 15 further steps by rolling out the prior with actions sampled from the actor (no observation correction). Value targets are $\lambda$-returns~\cite{schulman2015gae} ($\gamma{=}0.99$, $\lambda{=}0.95$) computed from the world model's reward head; the actor is updated by REINFORCE with advantage clipping, the critic by MSE, and entropy is decayed from $0.5$ to $0.15$ over training. We train for $1000$ epochs at learning rate $3{\times}10^{-5}$ and batch size $256$. This stage uses \emph{zero additional real-environment interactions}: all learning happens in the learned world model on the same offline dataset used to train it.

\paragraph{Model-free A2C (MF-AC).} Actor and critic are 3-layer MLPs of width $256$ over the raw observation, trained in the real environment for $1000$ epochs at $50$ episodes per epoch ($\approx{18.4\text{M}}$ transitions), with learning rate $3{\times}10^{-4}$, $\gamma{=}0.99$, $\lambda{=}0.95$, and entropy decayed from $0.2$ to $0.01$.

\paragraph{Behaviour cloning (BC).} A policy with the same architecture as the MF-AC actor is trained by cross-entropy on the human action, using only the $285$ good training episodes with return at least $+200$ (about $86$k transitions, part held out for early stopping). The checkpoint is chosen by validation cross-entropy, with no environment interaction, and training takes under a minute. At test time we evaluate both the argmax (deterministic) action and an action sampled from the predicted distribution (stochastic). BC does not use the world model.

\begin{table}[htbp]
\centering
\small
\begin{tabular}{@{}l >{\raggedright\arraybackslash}p{6.4cm} >{\raggedright\arraybackslash}p{2.9cm}@{}}
\toprule
\textbf{Tag} & \textbf{Training} & \textbf{Hardware} \\
\midrule
\texttt{12345-3080-R} & trained to epoch $300$, then resumed to $500$ in a new process$^{\ast}$ & RTX 3080 \\
\texttt{12345-3080-C} & one continuous process & RTX 3080 \\
\texttt{12345-5090-C} & one continuous process & RTX 5090 \\
\texttt{12345-5090-A} & as \texttt{12345-5090-C} to epoch $300$, then AdamW moments reset & RTX 5090 \\
\texttt{777-3080-C}   & one continuous process & RTX 3080 \\
\texttt{777-5090-A}   & as \texttt{777-3080-C} to epoch $300$, then AdamW moments reset & RTX 5090 after epoch $300$ \\
\texttt{777-L40S-C}   & one continuous process & L40S \\
\texttt{12345-3080-Z}, \texttt{12345-5090-Z} & one continuous process, reward head reads $\bm{z}$ only (Section~\ref{sec:zonly}) & RTX 3080 / 5090 \\
\bottomrule
\end{tabular}
\caption{Training runs referred to by tag. All use the human dataset and the configuration of this section; the first number is the training seed. $^{\ast}$The resume restarted the AdamW moments and re-seeded the random stream. Other seed panels (Sections~\ref{sec:domain} and~\ref{sec:heterogeneity}) are referred to by seed.}
\label{tab:runs}
\end{table}

\section{Model-Based A2C Across World-Model Checkpoints}
\label{sec:app-ac}

To see how imagination-trained policies depend on world-model quality, we trained one A2C agent on each of nine purposively sampled checkpoints of \texttt{12345-3080-R}, spanning early training, the pre-plateau, the MPC plateau and the smoothed-MPC oracle (WM~$60$ is a raw-MPC outlier whose $20$-episode mean spiked to $+173.5$ while the smoothed curve stayed below $+100$). All nine runs share the hyperparameters of Appendix~\ref{sec:app-model} ($1000$ epochs, a checkpoint every $50$ epochs), and every saved actor is evaluated on $100$ deterministic episodes. Table~\ref{tab:ac-summary} reports, per world model, the A2C checkpoint with the highest mean return (best-by-mean, BBM) and the one with the highest worst-case return (safest-by-worst, SBW), the more useful pick for deployment.

\begin{table*}[htbp]
\centering
\footnotesize
\setlength{\tabcolsep}{3pt}
\resizebox{\textwidth}{!}{%
\begin{tabular}{@{}r l c r r r r c r r r r c@{}}
\toprule
& & \multicolumn{5}{c}{\textbf{Best-by-mean A2C checkpoint}} & \multicolumn{5}{c}{\textbf{Safest-by-worst A2C checkpoint}} & \\
\cmidrule(lr){3-7}\cmidrule(lr){8-12}
\textbf{WM} & \textbf{MPC regime}
  & \textbf{Ep.} & \textbf{Mean} & \textbf{Worst} & \textbf{Catas} & \textbf{Perfect}
  & \textbf{Ep.} & \textbf{Mean} & \textbf{Worst} & \textbf{Catas} & \textbf{Perfect}
  & \textbf{\#safe} \\
\midrule
$50$  & early training       & $550$  & $+178.3$ & $-120.8$ & $5$ & $64$ & $900$  & $+141.3$ & $-74.7$  & $0$ & $47$ & $1/20$ \\
$60$  & raw-MPC outlier      & $500$  & $+206.3$ & $-157.3$ & $5$ & $79$ & $600$  & $+182.1$ & $-154.7$ & $8$ & $73$ & $0/20$ \\
$100$ & early training       & $800$  & $\mathbf{+223.3}$ & $-108.0$ & $1$ & $85$ & $200$  & $+65.0$  & $-80.2$  & $0$ & $25$ & $3/20$ \\
$200$ & pre-plateau          & $1000$ & $+191.4$ & $-184.2$ & $3$ & $70$ & $300$  & $+149.1$ & $-111.3$ & $1$ & $43$ & $0/20$ \\
$250$ & pre-plateau          & $300$  & $+141.0$ & $-161.2$ & $1$ & $26$ & $200$  & $+127.3$ & $-104.8$ & $1$ & $28$ & $0/20$ \\
$280$ & MPC plateau          & $800$  & $+217.5$ & $-178.9$ & $2$ & $79$ & $300$  & $+191.3$ & $\mathbf{-39.6}$  & $0$ & $\mathbf{71}$ & $2/20$ \\
$300$ & plateau midpoint     & $850$  & $+205.9$ & $\mathbf{-95.5}$  & $\mathbf{0}$ & $74$ & $600$  & $+138.3$ & $-63.1$  & $0$ & $49$ & $\mathbf{7/20}$ \\
$310$ & smoothed-MPC oracle  & $1000$ & $+209.6$ & $-131.0$ & $2$ & $75$ & $400$  & $\mathbf{+196.0}$ & $-40.4$  & $0$ & $52$ & $3/20$ \\
$315$ & MPC plateau          & $800$  & $+178.0$ & $-203.5$ & $1$ & $63$ & $700$  & $+133.1$ & $-72.1$  & $0$ & $47$ & $1/20$ \\
\bottomrule
\end{tabular}%
}
\caption{A2C training outcomes across purposively sampled world-model checkpoints of \texttt{12345-3080-R}, $100$ deterministic episodes per A2C checkpoint. \textbf{Ep.}\ is the AC training epoch of the reported checkpoint; \textbf{mean}/\textbf{worst} are mean and worst-case return; \textbf{catas} counts episodes with return $<-100$, \textbf{perfect} counts $>+200$; \textbf{\#safe} is the number of A2C checkpoints (out of $20$) with $0$ catastrophic episodes.}
\label{tab:ac-summary}
\end{table*}

A deployable world model is one whose BBM and SBW checkpoints both have a high mean. The plateau checkpoints satisfy this best: WM~$280$ (BBM $+217.5$, SBW $+191.3$) and the smoothed-MPC oracle WM~$310$ (BBM $+209.6$, SBW $+196.0$). WM~$100$ is the trap that the BBM column alone hides: it has the highest BBM ($+223.3$), but its safest checkpoint falls to $+65.0$.

\section{ROF Estimator and Reproducibility}
\label{sec:app-repro}

\paragraph{Estimator noise.} ROF estimates are noisy at the level of single checkpoints. Recomputing the same checkpoints with $64$ instead of $128$ Jacobian states leaves the mean unchanged to four decimals but gives a per-checkpoint correlation of only $r{=}0.50$, with a mean absolute difference ($0.009$) comparable to the entire within-run spread (s.d.\ $0.011$). Averaging over three metric seeds and a window of ${\sim}20$ checkpoints brings the standard error of a window mean to about $0.0008$, which is what makes the differences in Section~\ref{sec:validity} visible.

\paragraph{Cross-pipeline agreement.} The ROF drawdowns of \texttt{12345-3080-R} and \texttt{12345-3080-C} reproduce to three decimals between our implementation and an independent one ($0.0609$ vs.\ $0.061$, $0.0642$ vs.\ $0.064$), and logged epoch-5 metrics reproduce to four decimals across PyTorch $1.12$ and $2.11$. Closed-loop return, by contrast, depends on the full training path (Section~\ref{sec:heterogeneity}).

\section{ROF in Whitened Coordinates}
\label{sec:app-coords}

Under an invertible change of latent coordinates $\bm{s}' = T\bm{s}$, the linearization matrices become $TAT^{-1}$, $TB$, $CT^{-1}$ and $RT^{-1}$, so $\mathcal{O}_H' = \mathcal{O}_H T^{-1}$ and $\ker \mathcal{O}_H' = T \ker \mathcal{O}_H$. Theorem~\ref{thm:support} is unchanged, but ROF, a ratio of Euclidean norms, is not: with $C = [1, 0]$, $R = [1, 1]$ and $T = \mathrm{diag}(a, 1)$ it goes from $1/2$ to $1/(1+a^2)$.

For each checkpoint we estimate the latent covariance $\Sigma_s$ from $1024$ validation states, collected as the Jacobian states are (Section~\ref{sec:estimation}), and recompute ROF with $\mathcal{O}_H \Sigma_s^{1/2}$ and $\Sigma_s^{1/2}\bm{r}$ in place of $\mathcal{O}_H$ and $\bm{r}$, keeping the same $10^{-3}$ threshold. Since $\Sigma_s'^{1/2} = T\Sigma_s^{1/2}U$ with $U$ orthogonal, the whitened fraction is invariant to every invertible linear change of coordinates; it is the share of the linearized reward's variance over the validation states that lies in observable directions. On synthetic systems a random $T$ moved raw ROF from $0.849$ to $0.945$ and left the whitened value unchanged to six decimals. Computing the whitened fraction leaves every other logged quantity unchanged; the raw ROF values reproduce the original sweeps ($19$ of $7{,}200$ values differ, by at most $0.0002$).

\begin{table}[htbp]
\centering
\small
\begin{tabular}{@{}l l l@{}}
\toprule
\textbf{Comparison} & \textbf{Raw ROF} & \textbf{Whitened ROF} \\
\midrule
Bad-pool level, 12 runs: rank correlation with raw & $1$ & $-0.09$ \\
Halving $d_h$: bad-pool level offset & $+0.098$ & $+0.007$ \\
Halving $d_z$: bad-pool level offset & $+0.001$ & $+0.017$ \\
$\bm{z}$-only vs.\ control (3080), good pool, epochs $\ge 400$ & $0.285$ vs.\ $0.196$ & $0.881$ vs.\ $0.836$ \\
Within-run $\rho_s$(ROF, MPC), Reacher, 8 runs & $+0.26$ to $+0.87$ & $-0.73$ to $-0.40$ \\
Within-run $\rho_s$(ROF, MPC), LunarLander, 8 runs & $-0.55$ to $+0.32$ & $+0.03$ to $+0.77$ \\
\bottomrule
\end{tabular}
\caption{Raw and whitened ROF on the same checkpoints. Levels are seed-averaged means over epochs $\ge 50$; offsets are against the two standard runs on the same data. Within-run correlations use pool-averaged ROF against MA-7 MPC return. Raw values here come from a CPU recomputation and can differ in the third decimal from Table~\ref{tab:zonly}, which reports the original sweep.}
\label{tab:coords}
\end{table}

Whitening also changes the level. In all $22$ LunarLander runs we recomputed, whitened ROF lies between $0.85$ and $0.93$ in both pools, against $0.20$ to $0.37$ raw: most of the reward gradient's Euclidean energy outside the observable subspace sits in directions along which the validation states barely vary.

\section{Additional Probes of the Trained Model}
\label{sec:app-probes}

\paragraph{How the amortized posterior corrects errors.} Injecting state errors of matched magnitude and running the posterior on true observations, observable-direction errors \emph{persist} ($0.64$--$0.93$ of their initial norm after $10$ steps) while kernel-direction errors decay ($0.11$--$0.17$). This is the reverse of what a Kalman filter would do, and it follows from the architecture: the posterior resamples only $\bm{z}$, so error decay in $\bm{h}$ is governed by the trained recurrence, which contracts unused kernel directions and preserves the observable ones. Kalman-style intuitions about what imagination ``forgoes'' should therefore be applied to the $\bm{z}$-channel alone.
\section{Offline Metric Screen}
\label{sec:app-metrics}

The screen in Section~\ref{sec:negative-tests} uses the $27$ per-checkpoint metrics that were logged on every run, listed in Table~\ref{tab:screen-metrics}. Time-varying Jacobian variants (\texttt{jac\_dyn\_*}) were not computed on every run and are not in the screen. Each metric is seed-averaged and MA-7 smoothed over epochs $>50$, then reduced to four summaries: the mean of that curve, its last value, its maximum drawdown from a trailing peak, and its trend (last minus first). That gives the $27{\times}4{=}108$ candidates.

\begin{table}[htbp]
\centering
\small
\begin{tabular}{@{}l p{10.6cm}@{}}
\toprule
\textbf{Metric} & \textbf{What it measures} \\
\midrule
\multicolumn{2}{@{}l}{\emph{Validation loss (teacher-forced, posterior-corrected)}} \\
\texttt{val\_loss} & Weighted validation loss of~\eqref{eq:loss} \\
\texttt{val\_kl} & Posterior--prior KL \\
\texttt{val\_recon} & Physics $+$ contact reconstruction loss \\
\texttt{post\_obs\_rmse} & One-step posterior observation RMSE \\
\texttt{post\_rew\_rmse} & One-step posterior reward RMSE \\
\midrule
\multicolumn{2}{@{}l}{\emph{Open-loop rollout (5-step warm-up, then $H{=}25$ prior steps on logged actions)}} \\
\texttt{ol\_obs\_start} & Observation RMSE at the first open-loop step \\
\texttt{ol\_obs\_avg} & Observation RMSE averaged over the horizon \\
\texttt{ol\_obs\_max} & Worst-step observation RMSE along the horizon \\
\texttt{ol\_obs\_end} & Observation RMSE at step $H$ \\
\texttt{ol\_rew\_start} & Reward RMSE at the first open-loop step \\
\texttt{ol\_rew\_avg} & Reward RMSE averaged over the horizon \\
\texttt{ol\_rew\_max} & Worst-step reward RMSE along the horizon \\
\texttt{ol\_rew\_end} & Reward RMSE at step $H$ \\
\texttt{ol\_cumrew\_err} & RMSE of $\sum_{k=1}^{H}\hat r_{t+k}$ against the logged cumulative reward \\
\midrule
\multicolumn{2}{@{}l}{\emph{Fixed-Jacobian linearization (Section~\ref{sec:jacobian})}} \\
\texttt{jac\_spec\_radius} & Mean spectral radius of $A$ \\
\texttt{jac\_spec\_radius\_max} & Per-state maximum spectral radius of $A$ \\
\texttt{jac\_ctrl\_rank} & Effective rank of $\mathcal{C}_H$ (threshold $10^{-3}\sigma_{\max}$) \\
\texttt{jac\_ctrl\_cond} & Condition number of $\mathcal{C}_H$ at that rank \\
\texttt{jac\_obs\_rank} & Effective rank of $\mathcal{O}_H$ \\
\texttt{jac\_obs\_cond} & Condition number of $\mathcal{O}_H$ at that rank \\
\texttt{jac\_rcf} & Reward controllability fraction (good-pool) \\
\texttt{jac\_ocf} & Fraction of decoder energy in the controllable subspace \\
\texttt{jac\_rof} & Good-pool ROF \\
\texttt{jac\_rof\_bad} & Bad-pool ROF \\
\midrule
\multicolumn{2}{@{}l}{\emph{Finite-difference sensitivities}} \\
\texttt{emp\_C} & Mean pairwise next-state distance under different actions \\
\texttt{emp\_O} & Transition response to perturbations of $\bm{z}$ \\
\texttt{emp\_L} & Max local Lipschitz quotient of $f_\theta$ over sampled state pairs \\
\bottomrule
\end{tabular}
\caption{The $27$ offline metrics in the $108$-candidate screen. Each is reduced to mean, final, drawdown and trend as described in the text.}
\label{tab:screen-metrics}
\end{table}

\texttt{jac\_ocf} (OCF) is the decoder analogue of RCF: if $P_{\mathcal{C}}$ projects onto $\mathrm{range}(\mathcal{C}_H)$, then $\mathrm{OCF} = \norm{C\, P_{\mathcal{C}}}_F^2 / \norm{C}_F^2$, the fraction of the decoder Jacobian's energy that lies in the controllable subspace.

\end{document}